\documentclass[letterpaper, 10 pt, journal, twoside]{IEEEtran}
\usepackage[utf8]{inputenc}

\usepackage{url}
\usepackage{cite}
\usepackage{amsmath,amssymb,amsfonts}
\usepackage{textcomp}
\usepackage{lettrine}
\usepackage{threeparttable}
\usepackage{subcaption}
\usepackage[export]{adjustbox}
\usepackage{leftidx}
\usepackage{siunitx}
\usepackage{multirow}
\usepackage{booktabs}
\usepackage{pifont}
\usepackage{balance}
\usepackage{times}
\usepackage{siunitx}
\usepackage[dvipsnames, table]{xcolor}

\usepackage[mathscr]{euscript}
\usepackage{acronym}
\usepackage{todonotes}

\DeclareSymbolFont{rsfs}{U}{rsfs}{m}{n}
\DeclareSymbolFontAlphabet{\mathscrsfs}{rsfs}

\usepackage{hyperref}
\usepackage{algorithm2e}
\RestyleAlgo{ruled}
\usepackage{mathtools}
\usepackage{graphicx,booktabs}
\usepackage{soul}
\definecolor{colorFst}{RGB}{168,201,146}    
\definecolor{colorSnd}{RGB}{207,220,174}   
\definecolor{colorTrd}{RGB}{253,240,207}   
\newcommand{\fst}{\cellcolor{colorFst}\bf}   
\newcommand{\snd}{\cellcolor{colorSnd}}      
\newcommand{\trd}{\cellcolor{colorTrd}}      

\newcommand{\change}[2]{\textcolor{black}{#2}}
\newcommand{\add}[1]{\textcolor{black}{#1}}

\newcommand{\changetwo}[2]{\textcolor{black}{#2}}
\newcommand{\addtwo}[1]{\textcolor{black}{#1}}
\newcommand{\removetwo}[1]{}

\newcommand{\addthree}[1]{\textcolor{black}{#1}}

\newcommand{\fstLbl}[1]{\colorbox{colorFst}{\textbf{#1}}}
\newcommand{\sndLbl}[1]{\colorbox{colorSnd}{#1}}
\newcommand{\trdLbl}[1]{\colorbox{colorTrd}{#1}}
\newcommand{\cmark}{{\color{PineGreen}\checkmark}}
\newcommand{\xmark}{{\color{red}\ding{55}\ }}

\newcommand{\boldparagraph}[1]{\vspace{0.3em}\noindent{\bf #1} }

\acrodef{3DSG}{3D Scene Graph}
\acrodef{GNN}{Graph Neural Network}
\acrodef{GAT}{Graph Attention Network}
\acrodef{DCS}{Dynamic Covariance Scaling}
\acrodef{F3DSG}{Factorized \ac{3DSG}}

\begin{document}






\title{\LARGE \bf Generation of Uncertainty-Aware High-Level Spatial Concepts in Factorized 3D Scene Graphs via Graph Neural Networks}


\author{Jose Andres Millan-Romera$^{1}$,
Muhammad Shaheer$^{1,\dagger}$,
Miguel Fernandez-Cortizas$^{1,\dagger}$,\\
Martin R. Oswald$^{2}$, Holger Voos$^{1}$, and Jose Luis Sanchez-Lopez$^{1}$%
\thanks{$^{1}$Authors are with the Automation and Robotics Research Group, Interdisciplinary Centre for Security, Reliability and Trust (SnT), University of Luxembourg. Holger Voos is also associated with the Faculty of Science, Technology and Medicine, University of Luxembourg, Luxembourg.
\tt{\small{\{jose.millan, muhammad.shaheer, miguel.fernandez, holger.voos, joseluis.sanchezlopez\}}@uni.lu}}%
\thanks{$^{2}$Author is with the University of Amsterdam.
\tt{\small{m.r.oswald@uva.nl}}}
\thanks{$^\dagger$ These authors contributed equally.}
\thanks{*
This work was partially funded by the Fonds National de la Recherche of Luxembourg (FNR) under the projects 17097684/RoboSAUR and C22/IS/17387634/DEUS.}%
\thanks{*
For the purpose of Open Access, and in fulfillment of the obligations arising from the grant agreement, the authors have applied a Creative Commons Attribution 4.0 International (CC BY 4.0) license to any Author Accepted Manuscript version arising from this submission.}
}

\maketitle
\thispagestyle{empty}
\pagestyle{empty}

\begin{abstract} 
\label{abstract}
Enabling robots to autonomously discover \change{emergent}{high-level} spatial concepts (e.g., rooms \add{and walls}) from primitive geometric observations (e.g., planar surfaces) within 3D Scene Graphs is essential for robust indoor navigation and mapping. These graphs provide a hierarchical metric–semantic representation in which such concepts are organized. To further enhance graph-SLAM performance, Factorized 3D Scene Graphs incorporate these concepts as optimization factors that constrain relative geometry and enforce global consistency. However, both stages of this process remain largely manual: concepts are typically derived using hand-crafted, concept-specific heuristics, while factors and their covariances are likewise manually designed. This reliance on manual specification limits generalization across diverse environments and scalability to new concept classes.

\add{This paper presents a novel learning-based method that infers spatial concepts online from observed vertical planes and introduces them as optimizable factors within a SLAM backend, eliminating the need to handcraft concept generation, factor design, and covariance specification. We evaluate our approach in simulated environments with complex layouts, improving room detection by 20.7\% and trajectory estimation by 19.2\%\changetwo{, and further validate it on real construction sites, where}{. Validated on real construction sites,} room detection improves by 5.3\% and map matching accuracy by 3.8\%. \removetwo{Results confirm that learned factors can improve their handcrafted counterparts in SLAM systems and serve as a foundation for extending this approach to new spatial concepts.}}

\end{abstract}

\section{Introduction}
\label{introduction}

\setcounter{footnote}{2}
High-level spatial concepts (e.g., rooms \add{and walls}\footnote{\addthree{Wall refers to a physical boundary separating rooms, represented by two opposite vertical planes, as a base structural unit of the building layout.}}) \changetwo{inferred from observable geometric primitives}{which are not directly observable but inferred from observable geometric primitives} \add{(e.g., vertical planes)} enable robots to construct a richer and more human-interpretable understanding of their environment through \acp{3DSG}~\cite{3d_scene_graph}. This widely adopted representation encodes metric-semantic entities as nodes and their relationships as edges within a hierarchical graph, yet it does not capture the \changetwo{probabilistic dependencies among these entities}{structural constraints and uncertainty among entities across hierarchical layers}.
\begin{figure}[t]
    \centering
    \includegraphics[width=0.5\textwidth]{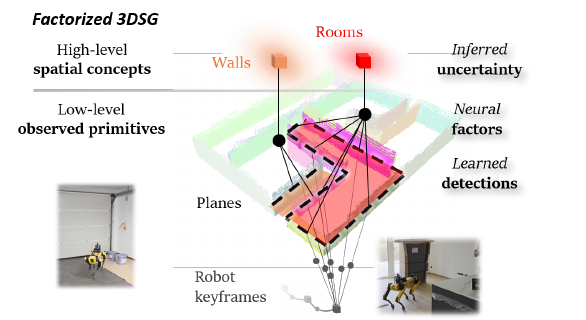}
    \caption{\addtwo{\textbf{Learned F3DSG.} We present a novel learning-based method to detect complex high-level spatial concepts (rooms and walls) modeled as learned factors and uncertainties.}}
    \label{fig:front}
\end{figure} 
\acp{F3DSG}~\cite{s_graphs+} address this limitation by \changetwo{explicitly}{coupling the \ac{3DSG} with the optimization backend, modeling high-level semantic entities as optimizable variables whose factors constrain the joint estimation of trajectory and map geometry within the factor graph.} Unlike approaches where such concepts exist only as post-processed scene descriptors ~\cite{hydra}, this formulation enables hierarchical dependencies (from low-level keyframes through geometric primitives to high-level concepts) to jointly refine both the robot trajectory and the scene representation, allowing semantic understanding to directly improve SLAM performance\changetwo{, which in turn enhances downstream tasks such as global localization~\cite{shaheer_deviations}, and path planning~\cite{ejaz2025situationally}.}{. This, in turn, enhances downstream tasks such as global localization~\cite{shaheer_deviations} or path planning~\cite{ejaz2025situationally}, which require semantic and geometric definitions.}



\addtwo{However, current methods for high-level spatial concept generation typically rely on concept-specific handcrafted algorithms. For example, for room detection, they depend on free-space clustering ~\cite{hydra}, occupancy-based segmentation \cite{werby2024hierarchical}, or require rectangular room configurations ~\cite{s_graphs+}. These geometric prerequisites limit applicability to layouts that satisfy them and do not generalize across concept types, as each demands a dedicated algorithmic design.}
Furthermore, although these methods construct 3DSGs with multiple semantic layers, they do not model high-level entities as optimization factors in the \changetwo{pose}{factor} graph.


\add{In \cite{s_graphs+}, the factorization of these high-level concepts was introduced, demonstrating their capability to improve the SLAM performance through joint optimization.} In this direction, \cite{reasoning_v1}, aim to generalize \add{the concept detection method} by learning pairwise relations among \textit{planes} with two models, one for \textit{walls} and one for rectangular \textit{rooms}. Nevertheless, in these methods, geometric factor definitions and their covariance remain handcrafted per concept and tied to the number of observed primitives, which ultimately hinders generalization to new \change{spatial emergent}{high-level spatial} concepts.

\add{This paper presents a novel learning-based method to generate online \change{spatial emergent}{high-level spatial} concepts of a \ac{F3DSG} as optimizable factors that are integrated into a SLAM backend. The presented work demonstrates this approach on rooms and walls conditioned on observed vertical planes, reducing the need to handcraft both the generation of these concepts along with their factor and covariance definition.} \changetwo{Our learning-based framework \textbf{(i)} detects the semantic-only 3DSG of these high-level spatial concepts with a per-concept confidence level, based on learned pairwise relations and temporal stabilization, \textbf{(ii)} learns to predict metric node attributes as centroids and their uncertainty for all generated nodes in the 3DSG, and \textbf{(iii)} defines learned factors with covariances computed from the semantic confidence and geometric prediction uncertainty, completing the F3DSG to be integrated in factor-graph SLAM.}{Our contributions are \textbf{(i)} a unified learning-based architecture that detects rooms and walls of higher complexity from pairwise plane relations, without requiring separate per-concept models, \textbf{(ii)} a learned metric inference module that predicts centroids for the generated high-level nodes (rooms and walls), eliminating the need to handcraft geometric factor definitions per concept type, and \textbf{(iii)} uncertainty-aware factor covariances estimation derived from both semantic detection confidence and geometric prediction uncertainty, replacing fixed handcrafted covariances and enabling the optimization backend to modulate the influence of each factor based on its reliability.}
%
%

We validate our method in a variety of indoor man-made environments, integrating them into state-of-the-art \ac{F3DSG}-based SLAM frameworks.

\section{Related work}
\label{related_work}

\subsection{Generation of high-level spatial concepts in 3DSGs}
\changetwo{The generation of \acp{3DSG} has progressed from merely placing observed objects and linking them by spatial proximity to modeling their semantic and geometric \emph{pairwise} relations ~\cite{wald2020learning}, and extended to open-world ~\cite{koch2024open3dsg}. They have excelled in applications like planning~\cite{gu2023conceptgraphs} or reasoning about functional relationships~\cite{zhang2025open}. 
However, these remain pairwise properties that can be resolved from the two connected nodes alone. More recent works go beyond pairwise relations to capture higher-order, set-level relations, such as belonging to the same room, which can only be determined by considering how multiple surrounding nodes are arranged together.
A prominent example is the notion of \emph{rooms} or \emph{spaces}, whose layouts were first inferred from 2D occupancy maps~\cite{rose2}. This concept was subsequently integrated into \acp{3DSG} by generating new high-level nodes and edges connecting to related low-level nodes via 3D free-space clustering~\cite{hydra,s_graphs+}, traversability~\cite{kim2025tacs} or similar concept-specific geometric methods for outdoor environments~\cite{greve2024collaborative}. These methods are largely ad hoc: they can excel at detecting specific set-level properties, but do not readily generalize to others. To overcome this limitation, other works classify observed objects into high-level nodes based on free-space, semantics, and relations between them. ~\cite{strader2024indoor} employs a learned ontology to group objects into higher-level nodes, while Clio~\cite{maggio2024clio} employs CLIP embeddings for task-tailored clustering. However, they do not generalize to structural elements of indoor environments, such as walls, and they are limited to semantic classification, not providing a metric definition of the created node.}{
Early \ac{3DSG} methods focused on modeling pairwise semantic and geometric relations among observed entities~\cite{wald2020learning}, and have since been extended to open-set settings~\cite{koch2024open3dsg}, excelling in applications like planning~\cite{gu2023conceptgraphs} and reasoning about functional relationships~\cite{zhang2025open}. However, these remain pairwise properties that can be resolved from the two connected nodes alone. More recent works go beyond pairwise relations to capture higher-order, set-level relations, such as belonging to the same room, which can only be determined by considering how multiple surrounding nodes are arranged together. Rooms were first introduced in \acp{3DSG} with hand-labeled annotations~\cite{3d_scene_graph} and later detected automatically via ESDF-based methods~\cite{rosinol20203d}, drawing on room segmentation techniques from occupancy representations~\cite{rose2, bormann2016room}. Subsequent works refined this through free-space clustering~\cite{hydra,s_graphs+}, occupancy-based segmentation~\cite{werby2024hierarchical}, or traversability~\cite{kim2025tacs}. These approaches are largely ad hoc and do not generalize across concept types. Other works group objects into higher-level nodes via learned ontologies~\cite{strader2024indoor} or CLIP-based clustering~\cite{maggio2024clio}, but do not extend to structural elements such as walls, nor produce geometric parameterizations suitable for optimization in the SLAM backend.}
\changetwo{On its side, graph completion aims to infer the unobserved part of a partially observed graph, including \emph{missing nodes} and their incident edges as a conditional reconstruction problem~\cite{kim2011network}. Deep, structure-conditioned generators explicitly \emph{grow} the graph by autoregressively adding new nodes and connections on top of a given conditioning subgraph~\cite{faez2022scgg}.
While most prior work targets synthetic graphs and molecules, robotics is increasingly adopting learning-based graph completion techniques for \acp{3DSG}. For instance, \cite{reasoning_v1} detects new \textit{wall} and \textit{room} nodes by classifying semantic relations among plane primitives, thereby extending the scene graph beyond observed elements; yet it employs a separate \ac{GAT}~\cite{velivckovic2017graph} per concept type, assumes rectangular layouts, and does not predict centroids for newly generated nodes.}{Complementary to detecting high-level concepts from geometric methods, learning-based approaches can infer new nodes in \acp{3DSG} by classifying relations among observed primitives. The authors of~\cite{reasoning_v1} detect wall and room nodes by classifying pairwise semantic relations among planes; yet this approach employs a separate \ac{GAT}~\cite{velivckovic2017graph} per concept type, assumes rectangular layouts, and does not predict centroids for the generated nodes.}

\addthree{Beyond detecting graph topology, \acp{GNN} can directly regress continuous node attributes conditioned on structure~\cite{scarselli2008graph} and neighboring attributes~\cite{fang2022invariant}, yet this capability has not been applied to high-level spatial concepts in \acp{3DSG}.}

\subsection{Learnable, Uncertainty-Aware Factor Graphs}

Factor graphs are the standard abstraction for large-scale probabilistic inference in SLAM, allowing heterogeneous constraints to fuse through sparse nonlinear least squares optimization \cite{dellaert2017factor}.  
Recent work couples scene graph structure to estimation, either by adding semantic priors to constrained factor graphs or by importing scene graph cues into the optimizer state \cite{haroon2024constrained}.  \textit{S-Graphs+}~\cite{s_graphs+, reasoning_v1} integrates a hierarchical \ac{3DSG} directly into the optimization state and improves trajectory and map accuracy, but defines each semantic concept (room–plane, etc.) with manually designed factors and fixed covariances, limiting extensibility.
In parallel, graph learning has begun to predict continuous factors with GNNs \cite{fang2022invariant}, and fully Bayesian graph convolutional networks estimate epistemic variance via Monte-Carlo dropout \cite{ryu2019bayesian}.  Several SLAM systems already modulate information matrices with external semantic-confidence scores, as in Probabilistic Data-Association SLAM \cite{bowman2017probabilistic} and \ac{DCS} \cite{agarwal2013robust}. \change{}{Exploring data-driven factors in graphs, \cite{moragrega2023data} demonstrates its use in anchor-based positioning via Bayesian inference.}
However, these techniques have not been applied to robotic \acp{F3DSG}, as current systems lack a unified and generalizable approach that learns and integrates geometric factors with uncertainty estimates capturing both semantic relation confidence and geometric prediction uncertainty for newly instantiated \change{emergent}{high-level} nodes. 
\section{Methodology}
\begin{figure*}[t!]
    \centering
    \includegraphics[width=0.89\textwidth]{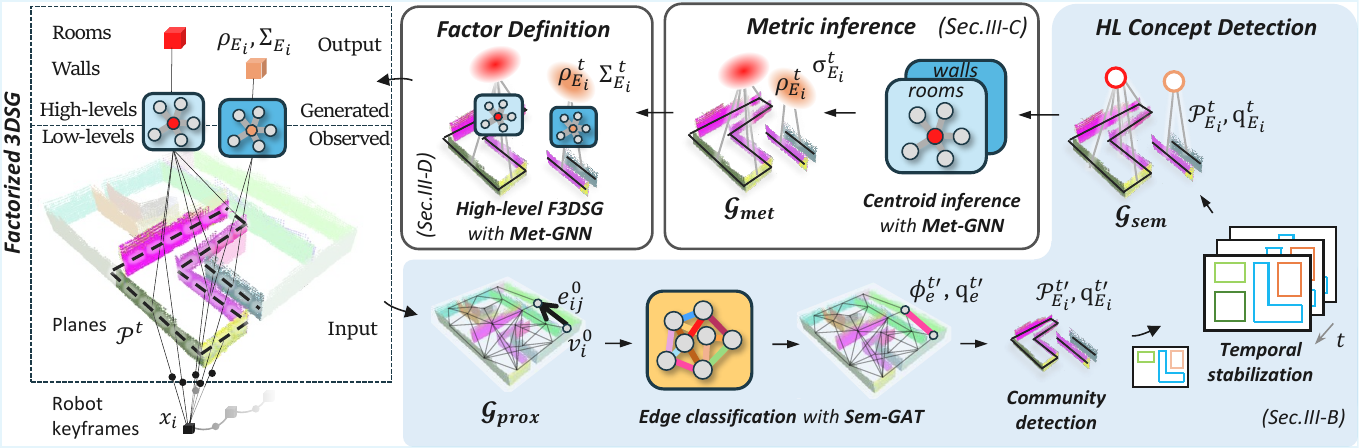}
    \caption{\textbf{System architecture.} From the plane layer of the SLAM backend, an initial graph by proximity is built connecting every node with its K neighbours. The Sem-GAT classifies edges into \textit{same room} or \textit{same wall}, which are separately clustered, generating a room or wall node per cluster if consistent with previous observations. The geometric origin of each new node is then inferred by its Met-GNN. A factor and its covariance are defined for every new node and incorporated into the \ac{F3DSG} for use by the SLAM backend.}
    \label{fig:system_architecture}
    \vspace{-0.4cm}
\end{figure*}
\label{methodology}
\subsection{Introduction}

\noindent\add{\boldparagraph{Background.} The SLAM backend incrementally constructs a factor graph composed of a \textit{keyframe layer} and a \textit{plane layer}. The \textit{keyframe layer} captures the robot trajectory, with consecutive keyframes connected through odometry factors. The \textit{plane layer} maintains persistent vertical planar landmarks extracted from LiDAR, where each plane $\pi$ is parameterized by its centroid, unit normal vector, and spatial extent. To improve robustness, noisy detections are rejected and overlapping planes are merged. The resulting plane observations $\mathcal{P}= \{\pi_0, \dots, \pi_n\}$ constitute the input to our method.}

\boldparagraph{System overview.} Each time the \textit{plane} layer of the \change{the F3DSG}{SLAM backend} is updated, our method infers a high-level graph and its connecting factors in three stages \addtwo{(Fig.~\ref{fig:system_architecture})}: (1) semantic \change{definition}{detection} groups planes into higher-level concepts such as rooms and walls using a \ac{GAT}\change{}{-based architecture}, \addtwo{with a temporal stabilization stage that tracks candidate communities across observations and only instantiates a node once consistent evidence has accumulated,} (2) metric definition uses a \ac{GNN} to predict centroids for the resulting high-level nodes, and (3) uncertainty-aware factor definition derives geometric constraints and covariances from the learned metric and semantic confidences for integration into the \ac{F3DSG}. \change{}{We employ two separate neural architectures to keep the semantic and metric components modular and to reuse the centroid predictor in factor construction.} The inferred room and wall nodes are added above the \textit{plane layer} and connected to their associated planes through factors, completing the \ac{F3DSG}.
\vspace{-0.1cm}
\subsection{High-Level Concept Detection}
\label{sec:semantic_gen}
\noindent\add{\boldparagraph{Initial graph.} After each plane set observation $\mathcal{P}^t$ at time $t$, we build a \emph{directed} proximity graph $\mathcal{G}_{\mathrm{prox}}$ by connecting every plane $\pi$ to its $k$ nearest neighbors (k-NN) by centroid-centroid Euclidean distance\addtwo{, where $i$ denotes the source and $j$ the target nodes}. The features are computed using normalized plane features in a 2D projection space. To maintain translation and rotation invariance, the initial node embedding $v_i^0$ is limited to its length, while edges encode local relative geometry: the Euclidean distance between centroids $d_{ij}$, the bearing between them $\varphi_{ij}$, and the relative angle between normals $\alpha_{ij}$. The bearing and the relative angle of the normals are expressed in the local-frame construction, yielding rotation and translation invariant descriptors, but are dependent on the reference frame of the $i$-th node, which is different to $j$-th node, imposing the need for a directed graph using a direction for each node.}
\addtwo{The definitions of the initial edge embeddings are defined in Eq.~\eqref{eq:initial_embeds}:
\begin{align}
\varphi_{ij} = \text{atan2}(p_j^y - p_i^y, p_j^x - p_i^x),
\alpha_{ij} = \cos^{-1}( \frac{\mathbf{n}i \cdot \mathbf{n}j}{|\mathbf{n}i||\mathbf{n}j|}) \label{eq:angles}\\
e{ij}^0 = [d{ij}, \varphi{ij}, \quad \alpha{ij}], e_{ji}^0 = [d_{ij}, \varphi{ji}, \alpha_{ji}] \label{eq:initial_embeds}
\end{align}}
\boldparagraph{Edge classification.} \changetwo{A \ac{GAT} (Sem-GAT) provides the probability $p_{ij,\theta}$ of each edge $e_{ij}$ to be classified as $\phi$:}{We employ a graph neural network with GAT-based message passing~\cite{velivckovic2017graph} (Sem-GAT) that follows an encode-process-decode structure to provide the probability $p_{ij,\theta}$ of each edge $e_{ij}$ to be classified as $\phi$:}
\begin{equation}
p_{ij,\theta}\!\big(\phi \mid e_{ij}, \mathcal{G}_{\mathrm{prox}}\big), \phi\in\{\textit{same-room},\textit{same-wall},\textit{none}\}
\label{eq:pairwise_prob}
\end{equation}
\changetwo{Via message passing, the encoder simultaneously updates node and edge embeddings \change{across $L{=}2$ hops}{}. Eq.~\eqref{eq:encoder_v} updates node embeddings via GAT layers \change{with $H{=}8$ attention heads}{} that are max-pooled and passed to an MLP $g_v$. Parallely, Eq.~\eqref{eq:encoder_e} updates edge embeddings via an MLP $g_e$,
where $v_i^l$ and $e_{ij}^{l}$ represent the node and edge embeddings at layer $l$ and $\mathcal{N}(i)$ is the vicinity of node $i$.
In Eq.~\eqref{eq:decoder}, node and edge embeddings of the last hop are jointly decoded with an MLP $g_d$ that outputs logits $c_{ij}$, followed by a softmax to obtain class probabilities.}{The encoding and message passing are unified: at each of $L$ hops, Eq.~\eqref{eq:encoder_v} updates node embeddings $v_i^l$ by applying a GAT layer followed by an MLP $g_v$, and Eq.~\eqref{eq:encoder_e} updates edge embeddings $e_{ij}^l$ via an MLP $g_e$, where $l$ denotes the hop and $\mathcal{N}(i)$ the vicinity of node $i$.}
\begin{align}
v_{i}^{l+1} &= g_{v}\!\Big([v_i^l, \max_{j\in\mathcal{N}(i)} \mathrm{GAT}_H(v_i^l, e_{ij}^l, v_j^l)]\Big) \label{eq:encoder_v}\\
e_{ij}^{l+1} &= g_{e}\!\Big([v_i^l, e_{ij}^l, v_j^l]\Big) \label{eq:encoder_e} \\
p_{ij} &=  \mathrm{softmax}\Big(g_d([v_i^L,e_{ij}^L,v_j^L]\Big) \label{eq:decoder}
\end{align}
\addtwo{These MLPs are integral to the message passing update at each hop, not applied between iterations. After the final hop, a decoder MLP $g_d$ (Eq.~\eqref{eq:decoder}) outputs logits $c_{ij}$, followed by a softmax, to obtain class probabilities.}
Training details in Sec.~\ref{training}.

Let $\phi_{ij}=\arg\max_\phi(p_{ij})$ be the predicted class. To compute epistemic uncertainty, we follow the Bayesian approximation of dropout proposed by~\cite{gal2016dropout}, applying dropout to all linear layers and keeping it active at test time. We then run\change{ $M{=}10$}{} stochastic forward passes per edge to define the semantic confidence $q^{t'}_{i,j}=\operatorname{Var}_{m=1}^{M}\!\big[\phi^{(m)}_{ij}\big]$.

To ensure robustness, only edges with $q^{t'}_{i,j}$ over a threshold $\tau_e$ are accepted as \textit{same-room} or \textit{same-wall}, otherwise they are set to \textit{none} class. Accounting for the case when both directions $i{\to}j$ and $j{\to}i$ are present, both are merged into a single edge $e$, rendering the graph undirected.
We set the class of the merged edge, $\phi^{t'}_e$ as the one with highest confidence, set as $q^{t'}_{e}$.

%
%
%
%
%
%
\boldparagraph{Community detection.} \change{Room communities are obtained by forming an undirected subgraph from edges classified as \textit{same-class} weighted by $q^{t'}_{e}$.
Greedy modularity maximization cite{clauset2004finding} receives this graph and yields the room communities $\{\mathcal{P}^{t'}_{R_i}\}$ and its confidence $q^{t'}_{R_i}$. Since walls only contain two planes, one community $\mathcal{P}^{t'}_{W_i}$ is defined for each \textit{same-wall} edge along with its associated $q^{t'}_{W_i}$. We generically denote them as $\mathcal{P}^{t'}_{E_i}$ and $q^{t'}_{E_i}$ for each community $E_i = \bigcup_{i} \{ R_i, W_i \}$.}{
To detect the existence of high-level entities (rooms and walls) $E_i = \bigcup_{i} \{ R_i, W_i \}$, we identify communities $\mathcal{P}^{t'}_{E_i}$ as subsets of planes in $\mathcal{P}^t$, whose internal edges are classified as \textit{same-entity}. In the case of rooms, communities are obtained by forming an undirected subgraph from edges classified as \textit{same-room} weighted by $q^{t'}_{e}$.
Greedy modularity maximization \cite{clauset2004finding} receives this graph and yields the room communities $\mathcal{P}^{t'}_{R_i}$ with a different weight $q^{t'}_{R_i}$ each. Since walls only contain two planes, one community $\mathcal{P}^{t'}_{W_i}$ is defined for each \textit{same-wall} edge along with its associated $q^{t'}_{W_i}$.}

\boldparagraph{Temporal stabilization.}
We account for room and wall perception changes that may occur when the geometry of planes is updated or when new planes are observed over the entire \textit{planes} layer. At $t{>}0$, current communities $\mathcal{P}^{t'}_{C}$ are matched to tracked communities $\mathcal{P}^{0:t-1}_{C}$ by the Community Matching $\mathcal{CM}$ module using intersection-over-union similarity $IoU(\cdot,\cdot)\ge\tau_J$ threshold with a hit/miss policy (initiate new tracks when unmatched; generate after $n$ similar observations; retire after $o$ misses).
\begin{equation}
\mathcal{P}^{t}_{E_i}=\mathcal{CM}(\mathcal{P}^{0:t-1}_{E_i},\mathcal{P}^{t'}_{E_i})
\label{eq:tracking}
\end{equation}
To stabilize decisions over time we use an \emph{exponential moving average (EMA)} of intra-communities confidence,
\begin{equation}
q^{t}_{E_i}=\beta\,q^{0:t-1}_{E_i} + (1{-}\beta)\,q^{\text{t'}}_{E_i}
\label{eq:ema}
\end{equation}
where $q^{0:t-1}_{E_i}$ is the confidence of the existing associated set $\mathcal{P}^{0:t-1}_{E_i}$,  $q^{t'}_{E_i}$ is the estimated confidence of the newly detected set $\mathcal{P}^{t^\prime}_{E_i}$, and $\beta\!\in\![0,1)$ controls smoothing. 
Finally, any set with $q^{t}_{E_i}$ over a threshold $\tau_{\mathrm{sn}}$ spawns a \textit{room} or \textit{wall} node, connected to its member planes to form $\mathcal{G}_{\mathrm{sem}}$.

\subsection{Metric Inference}
\label{sec:metric_gen}

We extend $\mathcal{G}_{\mathrm{sem}}$ by predicting the centroids of the \change{emergent}{high-level} nodes, yielding the metric–semantic graph $\mathcal{G}_{\mathrm{met}}$. 
We define a \textit{Met-GNN} architecture trained separately for each \change{emergent}{high-level spatial} concept type (See Sec.~\ref{training}).

For each community $E_i$ and its emerged node, we form an induced star graph with its incident \textit{plane} nodes $\mathcal{P}_{E_i}$. Plane features $\boldsymbol{\pi}_j$ are the centroid $\mathbf{p}$, unit normal $\mathbf{n}$, and length $\ell$ (normalized features), expressed in the current map frame; no edge features are used. To account for complex centroid definitions in the training dataset, Met-GNN performs a single-hop message passing with MLP $f^n_{\theta_C}$ from planes to the \change{emergent node}{high-level entity}, followed by mean pooling (permutation invariance over $\mathcal{P}_{E_i}$) and another MLP $f^e_{\theta_C}$ that outputs a 2-dimensional centroid, as defined in Eq.~\eqref{eq:room_centroid}:
\begin{equation}
\boldsymbol{\rho}_{E_i} \;=\; f^n_{\theta_C}\big(mean(f^e_{\theta_C}(\mathcal{P}^{t'}_{E_i})\big)),
\label{eq:room_centroid}
\end{equation}




To quantify epistemic uncertainty at time $t$, we apply dropout in all linear layers \change{and run $M{=}10$}{with} stochastic forward passes per \change{emergent node}{high-level entity}; the sample variance of the predicted centroid yields $\sigma^t_{E_i}$ (per-dimension), which are required for composing factor covariances.

\subsection{Factor Graph Definition}
\label{sec:factor_gen}

\boldparagraph{Factor definition.} We attach a \emph{single multi-plane factor} to each \change{emergent node}{high-level entity}, connecting its centroid variable to its supporting planes: an $N$-ary factor for a room (all planes in $\mathcal{P}_{R_i}$) and a binary factor for a wall (its two planes). These factors enforce geometric consistency via the learned predictors from Sec.~\ref{sec:metric_gen}.

For a node of the community $\mathcal{P}_{E_i}$ with previous centroid variable $\boldsymbol{\rho}'_{E_i}$ and plane variables $\{\boldsymbol{\pi}_j\in\mathcal{P}_{E_i}\}$, the costs are defined as:
\begin{equation}
\kappa_{E_i} \;=\; \boldsymbol{\rho}'_{E_i} \;-\; f_{\theta_R}\!(\mathcal{P}^{t'}_{E_i})
\label{eq:room_residual}
\end{equation}
%
Each factor contributes to the quadratic cost $\|\kappa\|^2_{\mathbf{\Sigma}^{-1}}$. Jacobians of all learned-factor residuals with respect to centroid and plane variables are computed by automatic differentiation in the optimizer.

\boldparagraph{Convariance definition.}Let $\sigma^{t}_{E_i}\in\mathbb{R}^2_{\ge 0}$ be the predictive variance from Met-GNN (Sec.~\ref{sec:metric_gen}) estimated via Monte-Carlo dropout samples for node $i$. We form the factor covariance $\mathbf{\Sigma}_i$ as a combination of $\sigma^{t}_{E_i}\in\mathbb{R}^2$ with the tracked community confidence $q^{t}_{E_i}$ via \ac{DCS}:
\begin{equation}
\mathbf{\Sigma}^t_{E_i} \;=\; \alpha\,\sigma^t_{E_i} \;\big/\, \max\!\big(q^{t}_{E_i},\,\varepsilon\big)
\label{eq:covariance_comp}
\end{equation}
with $\alpha,\varepsilon>0$ chosen on a validation set. This positive semidefinite form down-weights a factor (larger covariance) when geometric uncertainty is high or semantic confidence is low.

\boldparagraph{Factor integration.}
\change{We inject factors for all generated high-level spatial concepts into the F3DSG of the SLAM backend for joint optimization.}{We inject planes-to-room and planes-to-wall factors for all inferred high-level spatial concepts into the \ac{F3DSG} of the SLAM backend, enabling joint optimization of plane, room, and wall geometry together with the keyframes\addtwo{.}}
To maintain consistency with prior observations, we add factors for newly inferred concepts, update those whose supporting planes change (using the revised centroids), and remove factors for concepts no longer observed.
\section{Training}
\label{training}

\boldparagraph{Training datasets.} We employ a custom synthetic dataset generator and MSD public dataset~\cite{msd}, both mimicking common building layouts in the form of graphs. They include all the required nodes, edges, and their semantic and geometric definitions of the layers of the SLAM backend, excluding the keyframes. To make the synthetic dataset as realistic as possible, L-shaped \textit{rooms}, \textit{wall} thickness, \textit{wall} length, \textit{plane} dropout, and number of planes in a room can be tuned. Additionally, we post-process the data with noise in orientation and position for all geometric definitions. From these graphs, we extract subgraphs to form inputs and ground truth, enabling unlimited training samples with target features that complement the limited coverage of MSD.

\boldparagraph{Edge classification.} \removetwo{See Fig.3 left. }To train Sem-GNN, we generate a full synthetic layout containing all \textit{planes} along with the \textit{same room} and \textit{same wall} edges. The graph is used as ground truth for training, where we define the initial graph by proximity by connecting each node to its $k=15$ nearest neighbors. The optimization process is guided by the cross-entropy criterion, using the Adam optimizer. \change{}{Hyperparameter search selected a 2-hop message-passing backbone with 8 attention heads. The decoder $g_d$ uses hidden widths $[188,\,236]$, while the encoder MLPs $g_v$ and $g_e$ use 18 and 55 hidden channels, respectively. Training uses a learning rate of $1.56\times10^{-3}$ with batch size 128. Monte Carlo dropout is set to 10 passes.}




\boldparagraph{Origin inference.} \removetwo{See Fig.3 right. }Layouts of the MSD dataset are used to train the Met-GNNs by extracting subgraphs for each higher-level node (\textit{room} or \textit{wall}). In this dataset, the method to define the center is unknown. The high-level node, its adjacent \textit{plane} nodes, and the edges between them are used as input to the Met-GNN, with the ground truth origin for the high-level node removed. The optimization process is guided by the mean squared error (MSE) criterion, and the Adam optimizer is used. \change{}{Hyperparameter search for $\theta_C$ selected a single hidden layer of size 23 for $f^e_{\theta_C}$ and two hidden layers of sizes 20 and 15 for $f^n_{\theta_C}$. Training uses a learning rate of $3.50\times 10^{-5}$ with batch size 64. Monte Carlo dropout is set to 10 passes.}

\section{Evaluation}
\label{results}

\subsection{Evaluation methodology}

\boldparagraph{Baselines.}
As summarized in Tab.~\ref{tab:methods}, we compare the \ac{3DSG} generation against three baselines: free-space room detection in \textit{Hydra}~\cite{hydra}, free-space room detection in \textit{S-Graphs}~\cite{s_graphs+}, and learning-based wall and room detection in \textit{SghsGnn}~\cite{reasoning_v1}. 

SLAM performance is compared to \textit{S-Graphs} and \textit{SgrhsGnn}, while \textit{Hydra} is omitted as \change{emergent}{its high-level nodes do not define factors for SLAM optimization}.
We provide two ablations to our full method (\textit{Ours}). First, we only use semantic information to compute the covariance, setting $\sigma^t_{E_i} = \mathbf{1}_{\mathbb{R}^{2}}$ (\textit{Ours-SF}). \changetwo{Second, we substitute Met-GNN for a naive factor that averages centroids with the handcrafted factor and sets almost-zero covariance (\textit{Ours-NF})}{Second, we replace the learned factor with a handcrafted, classic factor (\textit{Ours-CF}), adapted from \textit{S-Graphs} to support an arbitrary number of planes, that computes the centroid as the average of its constituent planes and uses a fixed covariance $\sigma^t_{E_i} = 0.001 \, \mathbf{I}_{2}$ empirically optimized}. \addtwo{In the case of Ours, the learned covariance is escalated by $\alpha=0.0001$ based on hyperparameter tuning.} \add{To ensure comparability, all the methods use \textit{S-Graphs+}~\cite{s_graphs+} as the SLAM backend.
This backend had already demonstrated improvements over state-of-the-art LiDAR SLAM systems without high-level factors (e.g., FAST-LIO2\cite{xu2022fast}) in terms of trajectory error and map accuracy. Our evaluation thus isolates the contribution of different high-level concept generation methods within this shared backend.}


\begin{table}[!t]
\centering
\setlength{\arraycolsep}{6pt}
\renewcommand{\arraystretch}{1.0}
\begin{tabular}{|l|ccc|ccc|c|}
\hline
\multicolumn{1}{c}{} & \multicolumn{3}{c}{\textbf{Semantic detection}} & \multicolumn{3}{c}{\textbf{Factors}}\\ \hline 
\textbf{Method} & \textbf{Room} & \textbf{Wall} & \textbf{SU} & \textbf{SU} & \textbf{MU} & \textbf{Definition} \\ \hline 
Hydra~\cite{hydra} & Free Space & \xmark & \xmark & \xmark & \xmark & \xmark\\ 
S-Graphs~\cite{s_graphs+} & Free Space & \xmark & \xmark & \xmark & \xmark & Classic\\ 
SghsGnn~\cite{reasoning_v1} & GNN$^1$ & GNN$^1$ & \xmark & \xmark & \xmark & Classic\\ \hline
Ours-\changetwo{N}{C}F & GNN$^2$ & GNN$^2$ & \cmark & \xmark & \xmark & Classic$^3$\\ 
Ours-SF & GNN$^2$ & GNN$^2$ & \cmark & \cmark & \xmark & GNN$^4$\\ \hline
Ours & GNN$^2$ & GNN$^2$ & \cmark & \cmark & \cmark & GNN$^4$\\ \hline
\multicolumn{8}{l}{SU/MU: semantic/metric uncertainty.\quad $^1$~separate architectures,}\\
\multicolumn{8}{l}{\quad $^2$~joint architecture, \quad $^3$~average of centroids,\quad $^4$~learned architecture.}\\
\end{tabular}
\caption{Baseline and ablation comparison relative to the high-levels of the \ac{3DSG} and the factor graph.}
\label{tab:methods}
\end{table}

\boldparagraph{Evaluation environments.}
\add{Since public datasets do not capture the layout complexity and the ground truth of high‑level concepts, point clouds, and trajectories required to evaluate our high‑level concept detection and factor‑graph SLAM, we have recorded} 8 environments simulated in Gazebo with data acquired with a 3D LiDAR sensor (denoted as \change{\textit{SLE}}{\textit{S-}}) and 5 real-world environments \change{(\textit{RLE})}{{\textit{R-}}} recorded with a 3D LiDAR in a Boston Dynamics Spot\textsuperscript{\textregistered} in offices and houses in construction with visible walls, as exemplified in Fig.~\ref{fig:front}. 
As shown in Fig.~\ref{fig:generation_examples}, \change{\textit{SLE}}{\textit{S-}} covers edge cases with many non-rectangular and large rooms while \change{(\textit{RLE})}{{\textit{R-}}} includes real environments with at least one non-rectangular room, challenging the capabilities of existing baselines.

\boldparagraph{High-level 3DSG detection metrics.}
We define a metric to evaluate the quality of the set of planes associated with each \change{emergent}{high-level spatial} concept, quantifying both correct coverage and error rate, given their impact on the structural accuracy of the graph and subsequent SLAM performance. As the concepts evolve during exploration, the final snapshot is used once the entire environment has been explored.

Let $\mathcal{R}={r_j}$ be the set of ground-truth entities (exemplified in room for clarity) and $\hat{\mathcal{R}}={\hat{r}_i}$ the set of predicted entities of the same type. Each entity is represented by the set of planes it is connected to, denoted $\mathcal{P}(r_j)$ and $\mathcal{P}(\hat{r}_i)$, respectively. \add{The ground-truth associations between planes and entities are defined from a reference map of each scene. For each method, the predicted associations are visually verified against this ground truth. In the case of \textit{Hydra}, since it does not provide any plane-room association, each plane is visually associated with the closest free-space cluster within a range. Each predicted entity is then paired with the ground-truth entity that shares the most planes, forming a one-to-one assignment.} 

With the confusion matrix defined in Eq.~\eqref{eq:confusion}, precision, recall, and intersection over union (IoU) are similarly computed for rooms and wall entities.
\begin{equation}
\label{eq:confusion}
\begin{gathered}
TP = \mathcal{P}(\hat{r}) \cap \mathcal{P}(r), \\
FP = \mathcal{P}(\hat{r}) \setminus \mathcal{P}(r), \quad
FN = \mathcal{P}(r) \setminus \mathcal{P}(\hat{r})
\end{gathered}
\end{equation}
%

\addtwo{To assess the stability of concept detection over time, we define the community variance as the average number of membership changes per timestep, where the change at each timestep is the total number of plane additions and removals (symmetric difference) across all tracked communities.}

\boldparagraph{SLAM performance metrics.}
To assess the contribution of the factors, we report Average Trajectory Error (ATE) in S- as the ground truth trajectory is not available in R-. We also compute Map Matching Accuracy (MMA) in S- and R- as RMSE between ground truth and estimated point clouds with visually aligned origins for R-, where ground truth origin alignment is not available. We assess as well the required runtime and the number of variability of \change{emergent entities}{generated high-level nodes} over the number of plane nodes in $\mathcal{G}_{\mathrm{prox}}$.

\boldparagraph{Hardware.} All experiments were conducted on a laptop equipped with an Intel® Core™ i9-12900H processor (14 cores, 20 threads, base frequency 2.9 GHz, turbo up to 5.0 GHz), and an NVIDIA T600 Laptop GPU with 4 GB VRAM (driver 535.183.01, CUDA 12.2).

\subsection{Results and Discussion}

\begin{figure}[!htbp]
\centering
\includegraphics[width=0.5\textwidth]{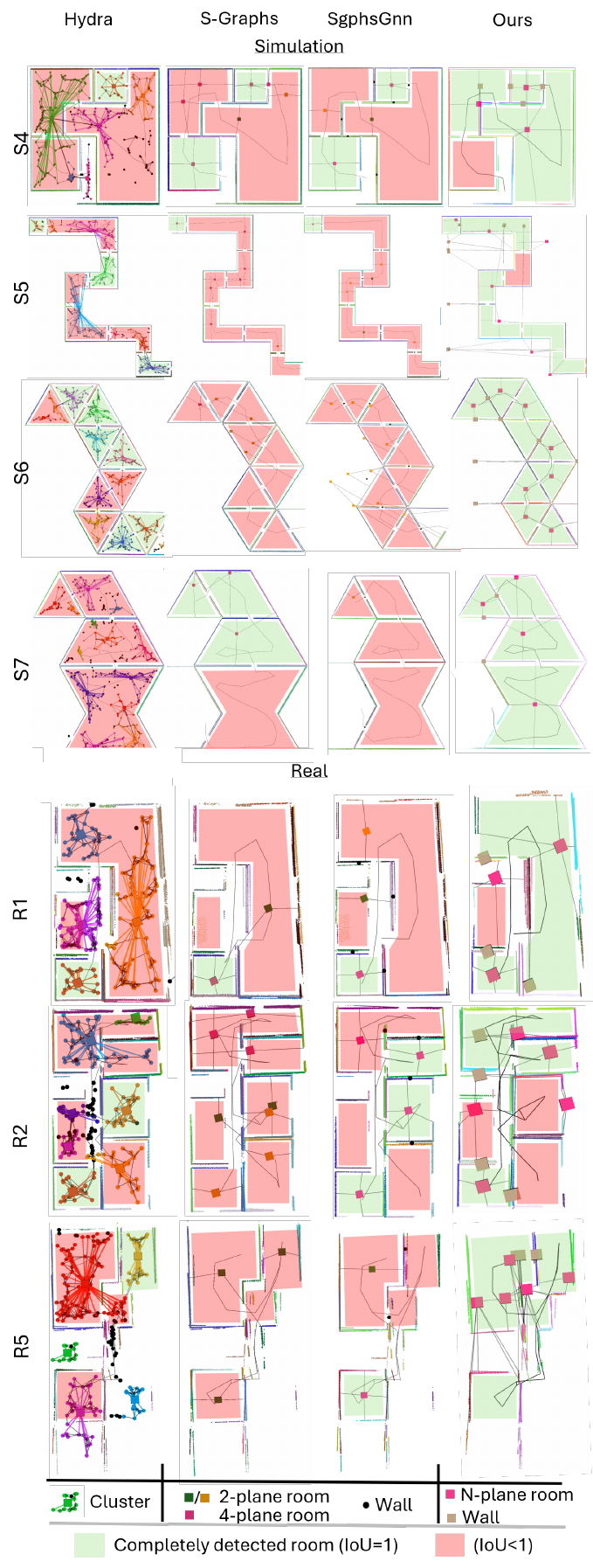}\\[-4pt]
\caption{\textbf{3DSG \change{generation}{detection} from LiDAR data.} Top-down views of RViz representation of the inferred 3DSGs. 
}
\label{fig:generation_examples}
\end{figure}

\begin{figure}[!htbp]
\centering
\includegraphics[width=0.5\textwidth]{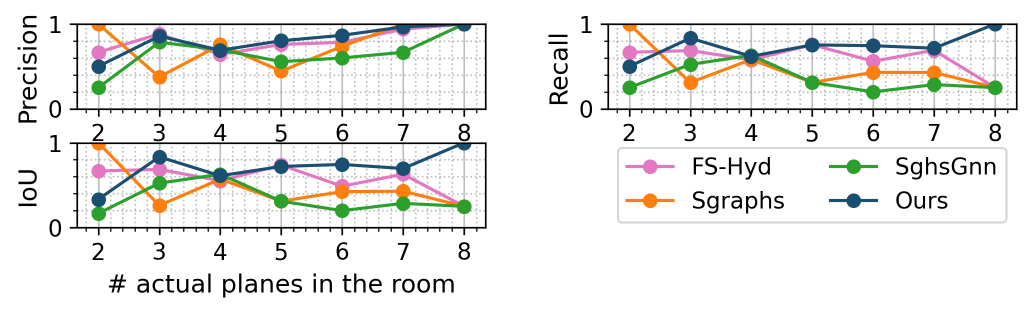}
\caption{\textbf{3DSG detection over room complexity.} Precision, recall, and intersection over union (IoU) performance of every detection method by number of planes.}
\label{fig:metrics_by_gt_planes}
\end{figure}

\boldparagraph{High-level 3DSG detection results.} 
Tab.~\ref{tab:rooms_allmetrics} shows consistent improvements in room detection across both simulation and real datasets: 3.4\%, 18.2\%, and 20.7\% (precision/recall/IoU) in S- and 9.5\%, 4.0\%, and 5.3\% on real data.
Fig.~\ref{fig:generation_examples} illustrates these improvements with representative layout examples and their detected entities.
The performance improvements are particularly notable in edge cases with complex, non-rectangular rooms, showing IoU gains of 44.5\% in S5, 30.8\% in S7, and 88.7\% in R1.
Fig.~\ref{fig:metrics_by_gt_planes} shows how this advantage is distributed across room complexity, particularly in rooms with 3 or 5+ \textit{planes} while maintaining state-of-the-art performance in rooms with 4 \textit{planes}.
In particular, L6 features only triangular rooms; S2, S3, S5, and all R- contain six-plane rooms; R4 contains an eight-plane room; and S7 includes non-Manhattan layouts.
Since \textit{S-Graphs} and \textit{SghsGnn} target simpler 2 or 4 plane configurations, they maintain the precision but decrease the recall, detecting correctly only a subset of actual planes.
\textit{Hydra} maintains comparable recall and IoU for larger rooms but fails on highly complex cases (8 planes or larger areas) in those challenging datasets. Our GAT-based approach successfully detects intricate room geometries that challenge rule-based methods, even though the training set only contains 2, 4, and 6-plane rooms, thanks to the out-of-distribution \change{generation}{detection} grounded on pair-wise relations and their clustering.

\change{For walls, Tab.~\ref{tab:walls_allmetrics}, demonstrate similar performance as \textit{SghsGnn} as 2-plane walls in non-rectangular layouts do not present differences.
Since walls are composed of two planes and both methods detect them either fully correctly or miss them entirely, the three metrics yield similar values; we therefore present only IoU.
Ours is 0.9\% better in S- and \textit{S-Graphs} 1.8\% better in R-, demonstrating state-of-the-art performance of our method.}{
Regarding walls, since they are composed of two planes and both methods detect them either fully correctly or miss them entirely, the three metrics yield similar values; we therefore present only IoU in Tab.~\ref{tab:walls_allmetrics}.
Ours presents 3.0\% higher IoU in S-, while the dedicated \ac{GNN} of \textit{SghsGnn} performs 1.8\% higher in R-, indicating that no method presents statistically significant improvements, and Ours matches the performance of state-of-the-art.}

\begin{figure}[!htbp]
\centering
\includegraphics[width=0.50\textwidth]{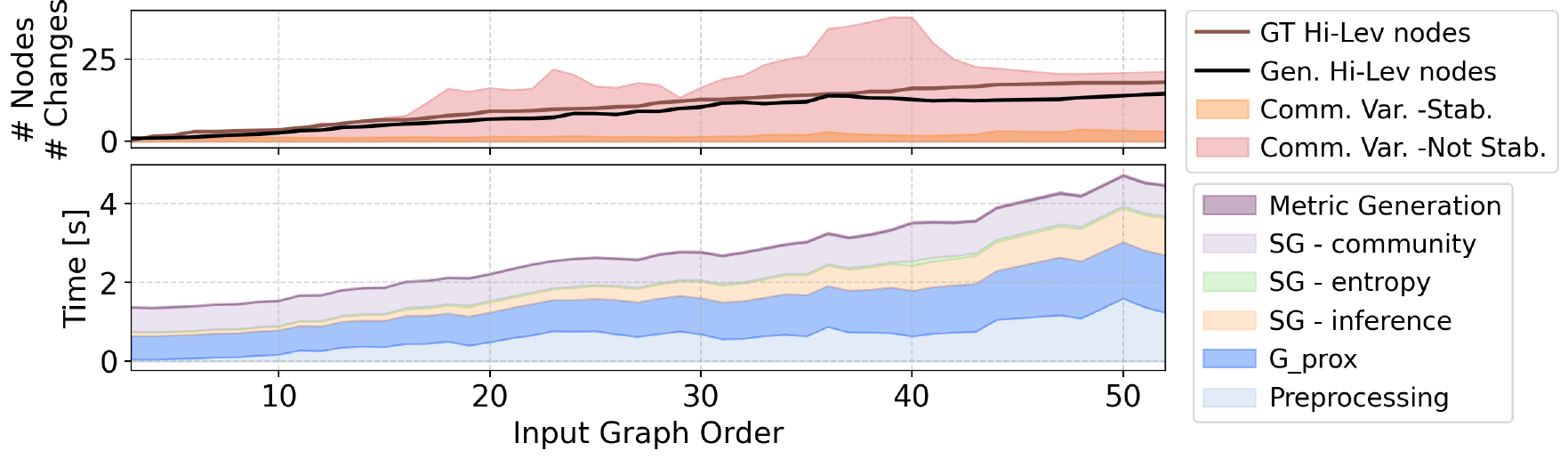}
\caption{\textbf{Computation times and inference variability.} (Top) Variability of community variables before and after stabilization, and number of generated high-level nodes. (Bottom) Per-stage computation time.}
\label{fig:avg_times_by_order}
\end{figure}

Fig.~\ref{fig:avg_times_by_order} (Top) presents how the number of instantiated concepts increases monotonically\changetwo{ while the variability of its associated plane communities}{, converging toward the ground-truth number of high-level nodes as exploration progresses. The community variance, measured as the variability of plane membership changes across timesteps,} is largely reduced by the \textit{Temporal stabilization} module, which enforces successive consistent detections before a concept is confirmed or updated.

\begin{table}[t]
\centering
 \setlength{\tabcolsep}{1.7pt} 
\renewcommand{\arraystretch}{1.05}
\scriptsize

\begin{tabular}{l|ccccccc|c|ccccc|c}
\toprule
\multicolumn{1}{c}{} & \multicolumn{8}{c}{\textbf{Simulation}} & \multicolumn{6}{c}{\textbf{Real}} \\
\cmidrule(lr){2-9}\cmidrule(lr){10-15}
Method & S1 & S2 & S3 & S4 & S5 & S6 & S7 & Avg & R1 & R2 & R3 & R4 & R5 & Avg \\ 
\midrule
\multicolumn{15}{l}{\textbf{Precision}} \\
\midrule
Hydra   & \snd 68.4 & \trd 66.7 & \trd 51.6 & \trd 78.9 & \snd 88.9 & \fst 100 & \snd 95.8 & \snd 78.8 & \trd 60.0 & \trd 70.8 & \trd 38.6 & \fst 65.0 & \fst 100 & \snd 66.9 \\
S-Graphs   & \trd 62.5 & \fst 85.7 & \fst 83.3 & \fst 81.2 & \snd 88.9 & \trd 45.0 & \trd 68.8 & \trd 74.4 & \snd 75.0 & \fst 86.1 & \fst 81.8 & 30.0 & \trd 41.7 & \trd 62.9 \\
SghsGnn  & \fst 85.7 & \trd 66.7 & 50.0 & \snd 80.0 & 77.8 & \fst 100 & 25.0 & 68.5 & \snd 75.0 & \snd 75.0 & \snd 66.7 & \trd 50.0 & 33.3 & 60.0 \\
\midrule
\textbf{Ours} & \fst 85.7 & \snd 80.5 & \snd 66.7 & 60.0 & \fst 100 & \snd 90.0 & \fst 100 & \fst 82.2 & \fst 100 & 64.3 & \snd 66.7 & \snd 60.3 & \snd 96.0 & \fst 77.4 \\
\midrule
\multicolumn{15}{l}{\textbf{Recall}} \\
\midrule
Hydra   & \trd 56.2 & \trd 66.7 & \snd 43.8 & 42.6 & \snd 59.3 & \snd 77.8 & \trd 57.3 & \snd 58.3 & \trd 36.0 & \fst 75.0 & 45.5 & \fst 66.7 & \snd 83.3 & \snd 61.3 \\
S-Graphs   & 53.0 & \snd 77.1 & \fst 66.7 & \snd 53.1 & \snd 59.3 & 33.3 & \snd 75.0 & \trd 56.1 & \snd 42.1 & 48.8 & \snd 54.5 & 20.0 & \trd 27.8 & 38.7 \\
SghsGnn  & \fst 77.1 & 53.3 & \trd 38.9 & \trd 51.7 & \trd 33.3 & \trd 66.7 & 16.7 & 47.7 & \snd 42.1 & \trd 57.1 & \fst 55.6 & \trd 46.2 & 22.2 & \trd 44.7 \\
\midrule
\textbf{Ours} & \snd 69.6 & \fst 78.8 & \fst 66.7 & \fst 60.0 & \fst 85.7 & \fst 90.0 & \fst 90.0 & \fst 76.5 & \fst 79.4 & \snd 60.2 & \trd 47.9 & \snd 54.4 & \fst 91.0 & \fst 66.6 \\
\midrule
\multicolumn{15}{l}{\textbf{IoU}} \\
\midrule
Hydra   & \trd 49.6 & \trd 66.7 & \trd 32.9 & 37.2 & \snd 59.3 & \snd 77.8 & \trd 56.7 & \snd 55.4 & \trd 36.0 & \fst 70.8 & 38.6 & \fst 65.0 & \snd 83.3 & \snd 58.8 \\
S-Graphs   & 48.8 & \fst 77.1 & \fst 66.7 & \snd 52.1 & \snd 59.3 & 29.0 & \snd 68.8 & \trd 54.1 & \snd 42.1 & 47.1 & \snd 54.5 & 20.0 & \trd 27.4 & 38.2 \\
SghsGnn  & \fst 77.1 & 53.3 & \snd 38.9 & \trd 51.7 & \trd 33.3 & \trd 66.7 & 16.7 & 47.5 & \snd 42.1 & \trd 57.1 & \fst 55.6 & \trd 44.4 & 22.2 & \trd 44.3 \\
\midrule
\textbf{Ours} & \snd 69.6 & \snd 75.6 & \fst 66.7 & \fst 60.0 & \fst 85.7 & \fst 90.0 & \fst 90.0 & \fst 76.1 & \fst 79.4 & \snd 57.8 & \trd 47.9 & \snd 50.5 & \fst 88.3 & \fst 64.8 \\
\bottomrule
\end{tabular}

\caption{Plane-level Precision/Recall/IoU (\%) for rooms across methods and LiDAR S- and R- datasets. Best results are highlighted by \fstLbl{first}, \sndLbl{second}, and \trdLbl{third}.}
\label{tab:rooms_allmetrics}
\end{table}


\begin{table}[t]
\centering
 \setlength{\tabcolsep}{1.7pt} 
\renewcommand{\arraystretch}{1.05}
\scriptsize
\begin{tabular}{l|ccccccc|c||ccccc|c}
\toprule
\multicolumn{1}{c}{} & \multicolumn{8}{c}{\textbf{Simulation}} & \multicolumn{6}{c}{\textbf{Real}} \\
\cmidrule(lr){2-9}\cmidrule(lr){10-15}
Method & S1 & S2 & S3 & S4 & S5 & S6 & S7 & Avg & R1 & R2 & R3 & R4 & R5 & Avg \\ 
\midrule
SghsGnn & \fst 100 & \snd 66.7 & \fst 85.7 & \fst 85.7 & \fst 100 & \snd  70.0 & \snd  66.7 & \fst 80.0 & \fst 100 & \fst 71.4 & \snd 75.0 & \snd  86.7 & \fst 66.7 & \fst 80.0 \\
\midrule
\textbf{Ours} & \fst 100 & \fst 75.0 & \snd 71.4 & \snd 71.4 & \snd 85.7 & \fst 77.8 & \fst 100 & \fst 83.0 & \snd 75.0 & \snd 66.7 & \fst 90.9 & \fst 91.7 & \fst 66.7 & \snd 78.2 \\
\bottomrule
\end{tabular}
\caption{Plane-level IoU (\%) for walls across methods and datasets. Best results are highlighted by \fstLbl{first} and \sndLbl{second}.}
\label{tab:walls_allmetrics}
\end{table}



\begin{figure}[!htbp]
\centering
\includegraphics[width=0.48\textwidth]{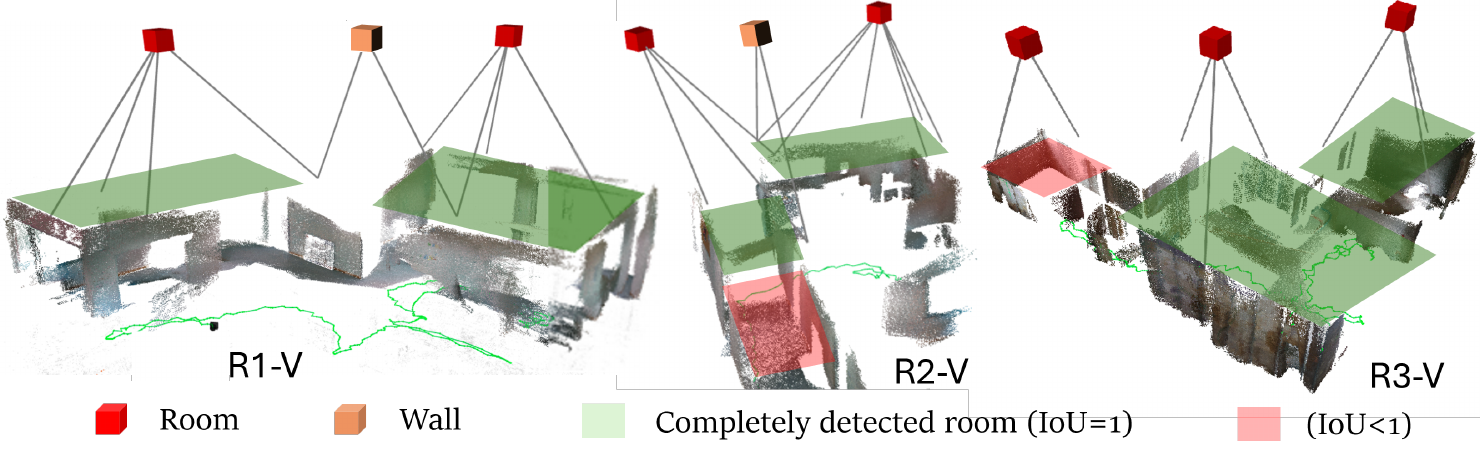}
\caption{\textbf{3DSG detection from RGB-D data.} Views of the \change{emergent}{high-level} room and wall nodes inferred from the point clouds of the planes. \removetwo{Rooms \textit{completely detected} (IoU=1) are depicted in GREEN while rooms with at least a wrong plane selection are depicted RED.}}
\label{fig:graph_generation_visual}
\vspace{1mm}
\end{figure}

\add{Additionally, we explore the applicability of our framework to different sensory inputs and SLAM backends by integrating it with visual S-Graphs\cite{tourani2025vs} on 3 real-world RGB-D environments captured with a handheld Intel RealSense D435. As shown in Fig.~\ref{fig:graph_generation_visual}, despite the different data distribution of the sensor, our method correctly detects most rooms and walls, indicating robustness to varying inputs and backend configurations.}



\boldparagraph{SLAM performance results.}

As shown in Tab.~\ref{tab:ate_average}, our full method reduces ATE by 19.2\% on average in S-, with the largest improvements of 53.8\% in \add{S5} and 36.8\% in \add{S7}, while maintaining performance on the remaining datasets.
\change{}{Note that S5 (non-rectangular) and S7 (non-Manhattan) are large-scale scenarios, where Ours improves room-IoU by +44\% and +31\%, respectively.}
This improvement stems from the unique exploitation of room-plane-wall-plane-room chains throughout the trajectory, which propagates geometric consistency and reduces drift in scenarios where loop closure cannot be applied.
\changetwo{Without the metric covariance (\textit{Ours-SF}), the improvement drops to 17.3\%, demonstrating the necessity to disable the factor when the uncertainty of the estimation is high}{Without the metric covariance (\textit{Ours-SF}), a similar error is achieved}.
The handcrafted geometric factors (\textit{Ours-\changetwo{N}{C}F}) achieve the largest improvement at 33\%, as their \changetwo{near-zero}{manually optimized factor definition and} covariance suits the low-noise simulated setting.


Regarding the MMA performance presented in Tab.~\ref{tab:mma_average}, the behaviour in S- is similar to ATE, notably improving for 46.5\% in \changetwo{L5}{S5} and 9\% in \changetwo{L7}{S7}, and maintaining the performance in the rest.
\changetwo{The best average improvement is provided by the hancoded, non-generalizable factors by 12.3\% (\textit{Ours-CF}). Our full learning-based factor provides an improvement of 11.0\% reduced to 10.8\% when not incorporating metric uncertainty into the covariance estimation (\textit{Ours-SF}).}{While \textit{Ours} and ablations present a similar MMA error, they reduce the it by 11.0\% on average with respect to the best baseline.}
When exposed to the real world in R-, our full method presents the largest improvement by 3.8\%, \changetwo{while the ablation of the metric uncertainty estimation reduces the improvement to 3.6\%, 
and the classic factor (\textit{Ours-CF}) achieves a comparable improvement, as the advantage of its handcrafted factor and covariance narrows under noisier real-world conditions. Conversely, the learned covariance is more conservative in simulation but better calibrated to real conditions, matching handcrafted performance without per-concept engineering.}{while the ablations present comparable reductions. However, missing the exploitation for the learned factor (\textit{Ours-SF}) incurs an increase of the standard deviation ($\pm$) of 47\% with respect to the full method.}
These ablation results demonstrate that learned inference with learned factors and covariance maintains \changetwo{SLAM performance}{the SLAM performance of ad-hoc, handcrafted factor definitions and tuned, fixed covariances.} Compared to the baselines, the proposed full method provides substantial improvements in edge cases with complex layouts \addtwo{without prior per-concept knowledge at a lower tuning effort}.


\begin{table}[t]
\centering
\setlength{\tabcolsep}{5pt}
\renewcommand{\arraystretch}{1.0}
\scriptsize
\newcommand{\val}[2]{\shortstack{#1\\[-2pt]{\tiny$\pm#2$}}}
\newcommand{\valn}[1]{\shortstack{#1\\[-3pt]{\tiny\phantom{$\pm0.00$}}}}
\begin{tabular}{lccccccc|c}
\toprule
Method & S1 & S3 & S4 & S5 & S6 & S7 & S8 & Overall \\
\midrule
S-Graphs & \snd\val{2.02}{0.01} & \snd\val{3.05}{0.01} & \val{3.40}{0.06} & \val{9.56}{0.02} & \snd\val{3.23}{0.01} & \val{4.83}{0.01} & \val{1.70}{0.00} & \val{3.97}{0.02} \\
SghsGnn  & \fst\val{2.00}{0.02} & \snd\val{2.64}{0.04} & \fst\val{3.05}{0.01} & \val{9.58}{0.03} & \trd\val{3.33}{0.01} & \trd\val{4.27}{0.10} & \val{1.78}{0.01} & \val{3.81}{0.03} \\
\midrule
\textbf{Ours} & \val{2.76}{0.10} & \trd\val{2.73}{0.03} & \trd\val{3.14}{0.30} & \trd\val{4.42}{0.48} & \val{3.39}{0.37} & \snd\val{3.54}{0.21} & \trd\val{1.56}{0.06} & \snd\val{3.08}{0.22} \\
Ours-\changetwo{N}{C}F       & \trd\val{2.19}{0.07} & \fst\val{2.20}{0.05} & \snd\val{3.09}{0.06} & \fst\val{3.33}{0.38} & \fst\val{2.47}{0.08} & \fst\val{3.27}{0.25} & \fst\val{1.26}{0.07} & \fst\val{2.55}{0.14} \\
Ours-SF       & \val{2.82}{0.08} & \val{2.87}{0.08} & \val{3.25}{0.08} & \snd\val{3.74}{0.47} & \val{3.52}{0.36} & \val{4.33}{0.06} & \snd\val{1.52}{0.11} & \trd\val{3.15}{0.18} \\
\bottomrule
\end{tabular}
\caption{\addtwo{\textbf{Average Trajectory Error (ATE)} [m $\times 10^{-2}$], of S-Graph+ with different detection modules on simulated environments from LiDAR. \fstLbl{Best}, \sndLbl{second}, and \trdLbl{third}. Standard deviation across environments is reported as $\pm$.}}
\label{tab:ate_average}
\end{table}
\begin{table}[t]
\centering
\setlength{\tabcolsep}{5.2pt}
\renewcommand{\arraystretch}{1.0}
\scriptsize
\newcommand{\val}[2]{\shortstack{#1\\[-2pt]{\tiny$\pm#2$}}}
\newcommand{\valn}[1]{\shortstack{#1\\[-3pt]{\tiny\phantom{$\pm0.00$}}}}
\begingroup
\begin{tabular}{lccccccc|c}
\toprule
\multicolumn{9}{c}{\textbf{Simulation}}  \\
\midrule
Method & S1 & S3 & S4 & S5 & S6 & S7 & S8 & Overall \\
\midrule
S-Graphs & \fst\val{26.36}{0.01} & \fst\valn{22.77} & \val{22.69}{0.00} & \val{39.88}{0.01} & \trd\val{22.11}{0.08} & \val{20.09}{0.01} & \trd\val{33.68}{0.02} & \val{26.80}{0.02} \\
SghsGnn  & \snd\val{26.40}{0.02} & \snd\val{22.86}{0.00} & \val{22.63}{0.04} & \val{41.03}{0.00} & \val{22.15}{0.29} & \val{20.08}{0.00} & \val{33.74}{0.02} & \val{26.98}{0.05} \\
\midrule
\textbf{Ours} & \val{26.53}{0.05} & \trd\val{23.02}{0.01} & \fst\val{22.36}{0.07} & \trd\val{21.32}{0.40} & \snd\val{21.90}{0.86} & \trd\val{18.27}{0.03} & \snd\val{33.53}{0.08} & \snd\val{23.85}{0.21} \\
Ours-\changetwo{N}{C}F       & \val{26.58}{0.03} & \val{23.19}{0.03} & \trd\val{22.48}{0.06} & \fst\val{19.27}{0.23} & \fst\val{21.70}{0.12} & \fst\val{17.83}{0.06} & \snd\val{33.53}{0.02} & \fst\val{23.51}{0.08} \\
Ours-SF       & \trd\val{26.47}{0.05} & \val{23.13}{0.07} & \snd\val{22.43}{0.02} & \snd\val{20.79}{0.56} & \val{22.70}{0.29} & \snd\val{18.22}{0.11} & \fst\val{33.47}{0.01} & \trd\val{23.89}{0.16} \\
\bottomrule
\end{tabular}

\medskip
\setlength{\tabcolsep}{8.4pt}
\begin{tabular}{lccccc|c}
\toprule
\multicolumn{7}{c}{\textbf{Real}}  \\
\midrule
Method & R1 & R2 & R3 & R4 & R5 & Overall \\
\midrule
S-Graphs & \val{76.71}{0.00} & \trd\val{67.15}{0.20} & \val{260.54}{0.12} & \snd\val{33.00}{0.61} & \snd\val{184.91}{0.02} & \val{124.46}{0.19} \\
SghsGnn  & \val{77.31}{0.00} & \val{67.45}{0.01} & \val{258.06}{1.02} & \val{34.27}{0.31} & \fst\val{184.79}{0.00} & \val{124.38}{0.27} \\
\midrule
\textbf{Ours} & \snd\val{76.02}{0.61} & \fst\val{65.37}{0.30} & \snd\val{235.70}{2.04} & \trd\val{33.68}{0.94} & \val{187.57}{0.15} & \fst\val{119.67}{0.81} \\
Ours-\changetwo{N}{C}F       & \trd\val{76.70}{0.12} & \val{68.25}{0.16} & \trd\val{235.87}{0.85} & \fst\val{31.76}{0.48} & \trd\val{187.38}{0.03} & \trd\val{119.99}{0.33} \\
Ours-SF       & \fst\val{75.81}{0.27} & \snd\val{65.85}{0.30} & \fst\val{234.12}{1.97} & \val{36.66}{3.55} & \val{187.40}{0.08} & \snd\val{119.97}{1.24} \\
\bottomrule
\end{tabular}
\endgroup
\caption{\addtwo{\textbf{Map Matching Accuracy (MMA)} [m $\times 10^{-2}$], of S-Graph+ with different detection modules on simulated (S-) and real (R-) data. \fstLbl{Best}, \sndLbl{second}, and \trdLbl{third}. Standard deviation across environments is reported as $\pm$.}}
\label{tab:mma_average}
\end{table}

\boldparagraph{Computation time.}
Fig.~\ref{fig:avg_times_by_order} (Bottom) shows that joint semantic and metric inference times increase monotonically with graph order yet remain within a few seconds, sustaining online operation in parallel with the SLAM back end. While preprocessing scales with order, $\mathcal{G}_{\mathrm{prox}}$ construction and community detection are only marginally affected by it, and entropy computations and metric generation remain negligible. The average computation time of a single factor is $691\,\mu s$, with the full optimization taking $2\,s$ in an environment with 40 planes. \addtwo{Hence, the overall computational cost remains compatible with online applications.}

\section{Conclusion}


This paper presents a learning-based method that generates high-level spatial concepts (rooms and walls) from observed vertical planes and instantiates them as optimization factors within \acp{F3DSG}. This removes the need to handcraft concept generation, factor definition, and covariance formulation\addtwo{, at a computational cost that remains compatible with online operation.}

\add{Across complex environments, our approach outperforms room detection baselines, particularly in non-rectangular layouts, while matching state-of-the-art wall detection. Our method yields the strongest SLAM gains in edge cases, though improvements decrease in real environments where predominantly rectangular layouts limit the advantage. In simulation, the fully handcrafted factor and covariance achieve the best ATE, yet both methods converge to equivalent performance in real data, suggesting that refining the learned architecture or training data could close this gap.}

Beyond the analyzed performance, the learning-based components are not inherently tied to rooms and walls conditioned on vertical planes. \changetwo{Extending to new high-level spatial concept types requires retraining with appropriate datasets and adapting community detection, but not redesigning factor or covariance formulations, unlike handcrafted approaches, where each concept demands a dedicated geometric design.}{Extending to new high-level spatial concept types (e.g. doorway from planes), potentially conditioned on different input primitives and producing different output entities (e.g. floor from rooms), requires adapting the input and output feature dimensions, the community detection strategy, and retraining with appropriate datasets, but the overall framework, including the learned factor and covariance formulation, remains unchanged without requiring dedicated geometric design per concept.}

\bibliographystyle{IEEEtran}
\bibliography{Biobliography}

\end{document}